\documentclass[conference]{IEEEtran}
\usepackage{times}
\usepackage{geometry}
\usepackage{graphicx}
\usepackage{amsmath}
\usepackage{hyperref}
\usepackage{tabularx}
\usepackage{booktabs}
\usepackage{amsfonts}

\title{Integrating Dynamical Systems Learning with Foundational Models: A Meta-Evolutionary AI Framework for Clinical Trials}

\author{
\IEEEauthorblockN{
Joseph Geraci$^{1,2,3,4}$, 
Bessi Qorri$^1$, 
Christian Cumbaa$^1$, 
Mike Tsay$^1$, 
Paul Leonczyk$^1$, 
Luca Pani$^{5,6}$
}

\IEEEauthorblockA{
$^1$NetraMark Corp., Toronto ON, Canada \\
$^2$Department of Pathology and Molecular Medicine, Queen’s University, Kingston ON, Canada \\
$^3$Tandem Centre for Pharmacogenetics, CAMH, Toronto ON, Canada \\
$^4$Arthur C. Clarke Centre for Human Imagination, UC San Diego, San Diego CA, USA \\
$^5$Leonard M. Miller School of Medicine, University of Miami, Coral Gables FL, USA \\
$^6$Department of Biomedical, Metabolic, and Neural Sciences, University of Modena and Reggio Emilia, Modena, Italy
}
}

\begin{document}
\maketitle
\vspace{2ex} 
\begin{quote}
\textit{``Maths has been an incredible tool for describing physics. In the same way I think AI might be an incredible descriptive language for biology.'' – Demis Hassabis (2024 Nobel Lecture)$^1$}
\end{quote}

\begin{abstract}
Artificial intelligence (AI) has evolved into an ecosystem of specialized ``species,'' each optimized for distinct strengths. We analyze two such systems: DeepSeek-V3, a 671-billion-parameter Mixture of Experts large language model (LLM) exemplifying scale-driven generality, and NetraAI, a dynamical system-based framework engineered for stability and interpretability on small clinical trial datasets. We formalize NetraAI’s foundations, combining contraction mappings, information geometry, and evolutionary algorithms to identify predictive patient cohorts. Patient features are embedded in a metric space and iteratively contracted toward stable attractors that define latent subgroups. A pseudo-temporal embedding and long-range memory mechanism enable exploration of higher-order feature interactions, while an internal evolutionary loop selects compact, explainable 2–4-variable bundles to define “Personas.”

To guide discovery, we introduce an LLM Strategist as a meta-evolutionary layer that observes NetraAI’s Persona outputs, prioritizes promising variables, injects domain knowledge, and assesses robustness. This two-tier architecture mirrors the human scientific process—NetraAI as experimentalist, the LLM as theorist—forming a self-improving loop. In case studies (schizophrenia, depression, pancreatic cancer), we compare standard classifiers, LLMs, and NetraAI, showing how NetraAI uncovered small, high-effect-size subpopulations that transformed weak baseline models (AUC \( \approx 0.50–0.68 \)) into near-perfect classifiers using only a handful of features. We position NetraAI at the intersection of dynamical systems, information geometry, and evolutionary learning, aligned with emerging concept-level reasoning paradigms such as LeCun’s Joint Embedding Predictive Architecture (JEPA). By prioritizing reliable, explainable knowledge, NetraAI offers a new generation of adaptive, self-reflective AI to accelerate clinical discovery.
\end{abstract}

\section{Introduction}

Advances in artificial intelligence (AI) have produced a rich taxonomy of approaches, often likened to an evolutionary branching of ``AI species.'' Early rule-based symbolic AI systems gave way to data-driven learners and ensemble methods for structured data, culminating in today’s massive deep neural networks (DNNs) capable of learning from both structured and unstructured data. More recently, large language models (LLMs), such as GPT-4 and its open-source counterparts, have dominated many benchmarks by scale alone—analogous to apex predators in a data-rich environment. For example, DeepSeek-V3, an open-source Mixture-of-Experts (MoE) Transformer, illustrates the extreme end of this evolution, consisting of 671 billion parameters (with approximately 37 billion active per token) and trained on 14.8 trillion tokens. This massive ``diet'' of data and compute yields broad capabilities with remarkable generality, but it comes at the cost of interpretability and data efficiency, rendering these complex black-box models ill-suited for biomedicine, where clinical datasets are small, sensitive, and high-dimensional, and stakeholders (i.e., clinicians and regulators) require transparent rationales for any AI-derived recommendations.

These limitations have catalyzed the emergence of a very different AI species—one that prioritizes stable dynamics, intrinsic interpretability, and domain knowledge integration over brute-force pattern recognition. In this work, we investigate NetraAI, a dynamical-systems-based framework specifically developed for clinical trial analysis, positioning it as a distinct species in the AI evolution landscape. By making choices that trade raw predictive coverage for reliability and clarity of insight, NetraAI addresses the niche of high-dimensional, small sample, high-stakes problems. As an embodiment of AI’s potential to become a ``descriptive language for biology,'' NetraAI builds explainability into its core. Its Persona-based approach reflects latent structures identified through outcome-guided convergence that must survive internal validation. Each Persona is a statistically and clinically meaningful patient subgroup defined by a compact (2–4) variables that consistently co-occur with a distinct clinical outcome, yielding directly actionable trial enrichment strategies. Notably, NetraAI’s Persona-based approach aligns with emerging trends in AI emphasizing concept-level reasoning. For instance, LeCun’s Joint Embedding Predictive Architecture (JEPA) trains models to predict in an abstract latent space instead of generating token-by-token. Similarly, NetraAI operates on high-level representations (Personas) rather than on raw textual output, marking a shift away from shallow pattern completion and toward embedding-based prediction.

Our goal here is to juxtapose two AI lineages—scale-driven LLMs (DeepSeek-V3) and dynamical-systems AI (NetraAI)—and explore how they might co-evolve. We begin with NetraAI’s theoretical roots: its metric-space embedding of clinical data, use of Banach contraction mappings to guarantee convergence, and its built-in long-range memory to tame combinatorial feature search. We then detail NetraAI’s two-tier evolutionary workflow: an inner genetic algorithm-like iterative refinement of transparent, predictive cohorts, and an outer LLM ``Strategist'' that orchestrates successive runs, injecting domain knowledge, monitoring uncertainty, and fusing discoveries—a process we term Meta-Evolutionary Design. Each step embeds verification and validation principles from reliability engineering, ensuring traceable and trustworthy outputs. By contrasting this cooperative division of labor with monolithic scaling, we argue that future AI evolution will be symbiotic, with LLMs acting as intelligent conductors orchestrating specialized scientific learners rather than outright competitors.

\section{Theoretical Framework}

\subsection{Attractor Dynamical Systems and Contraction Mappings}

At the core of NetraAI is the idea that learning can be formulated as a dynamical system whose iterative updates converge to stable fixed points—attractors—in a high-dimensional patient feature space. Specifically, NetraAI applies an iterative transformation map $F$, constructed as a contraction on a well-defined metric space, directly to the data representation space, such that it maps related objects closer together at each iteration. By the Banach Fixed-Point Theorem, any contraction mapping guarantees convergence to a unique fixed point. NetraAI represents each patient as a vector in feature space, and successive applications of $F$ progressively draw together patients whose latent data share similar, outcome-relevant patterns, causing them to fall into the same basin of attraction—a stable cohort reflecting a latent clinical subgroup. Once the system enters a region of meaningful structure, it converges reliably to a stable state. This mathematical foundation distinguishes NetraAI from conventional ML by providing convergence guarantees and consistency in the discovery of high-impact subgroups.

This contraction-based approach confers several key advantages over conventional ML:

\begin{enumerate}
    \item \textbf{Global Stability and Robustness}: Small perturbations or differences in initialization do not lead to divergent results and instead are dampened as the system's dynamics evolve toward attractors. Each attractor represents a latent cohort that persists across runs and is defined by a shared relationship to the target outcome.
    \item \textbf{Implicit Long-Range Memory and Noise-Filtering}: Each iteration of $F$ aggregates information across all patients, smoothing idiosyncratic noise while reinforcing stable, outcome-relevant patterns—akin to attention mechanisms in neural networks that highlight persistent features, but here emerging naturally from the contractive dynamics that emphasize variables and substructures that persist over iterations and contribute to outcome separation.
    \item \textbf{Mesoscopic Insight via Early Stopping}: Rather than running the dynamical system to full convergence (where all points collapse into a single attractor), NetraAI halts when a chosen purity or information criterion peaks. This captures mesoscopic configurations—intermediate-scale groupings that are more nuanced than broad clusters, yet more robust than transient noise. Such configurations correspond to explainable, clinically meaningful patient subpopulations defined by just a few high-impact variables.
\end{enumerate}

In effect, NetraAI transforms learning into a pseudo-temporal exploration of the clinical data’s geometry; each iteration reveals new candidate signals as patients regroup according to outcome-relevant features through supervision and loss function scoring. The contraction framework guarantees that once meaningful groupings emerge, they will be reinforced and not lost to noise. By halting at the point of maximal structure, NetraAI isolates clear, high-effect subgroups that would be obscured by either a single global model or by running to full convergence, where ultimately all points merge. The result is an intrinsically interpretable embedding of the data: patient positions in the latent space reflect clinically significant similarities, free from the arbitrariness of one-shot clustering or the opacity of large black-box models.

\subsection{Pseudo-Temporal Embedding and Higher-Order Feature Interactions}

A distinctive aspect of NetraAI is that the embedding process itself is data-driven and sequential – constructed through the pseudo-time evolution of the dynamical system rather than real chronology. Concretely, each feature and its value act as a small transformation on the latent state, nudging patient representations closer or farther apart depending on that feature’s contribution to outcome separation. The ordering of variables defines a path through pseudo-time, an ordered sequence of feature influences. As the system evolves through this sequence, an adaptive embedding emerges: patients’ latent coordinates are shaped not by just individual features, but by how variables interact to pull them toward distinct outcomes. This approach treats the variables themselves as parameters of the dynamical system’s evolution, effectively making the feature set an integral part of the model’s parameter space.

This ordered process systematically explores higher-order feature interactions by varying feature sequences and monitoring the resulting convergence behavior. NetraAI can adaptively eliminate (skip or down-weight) features that do not improve outcome separation and reinforce those that do, through an elimination reinforcement process. When a combination of features yields a pronounced cluster purity increase (i.e., how well patients separate into different outcome groups) through NetraAI’s long-range memory mechanism, the system retains these variables, preserving what it learns across runs. This ensures that true high-impact feature bundles are discovered and maintained as the system evolves. This strategy circumvents exhaustive subset search in an astronomically large feature space by reducing the computational complexity, while having the ability to approximately explore this space efficiently.\textsuperscript{6}

By elevating features to first-class parameters in the learning process, NetraAI uncovers interaction patterns often invisible to conventional techniques. Similar to how Nicolau et al., identified a novel breast cancer subtype using topological data analysis, NetraAI aims to focus on interactions to automate multi-factor profile discovery.\textsuperscript{7} When this learning strategy is applied to discover clinical trial enrichment criteria, NetraAI enforces sparsity for interpretability and feasibility by inherently limiting itself to a small variable limit for each Persona, avoiding the pitfall of overfitting to spurious complexities. This dynamic approach reveals new taxonomies of disease as the data speaks for itself beyond the dependent supervising variables, particularly important in medical sciences where human-derived labels of disease response vastly underrepresent the complexity and heterogeneity of patient populations.

\subsection{Long-Range Memory for Feature Synergy Discovery}

Biologically, memory enables an organism to integrate information over time — analogously, NetraAI’s long-range memory is woven directly into its contraction dynamics, enabling the algorithm to integrate information over successive feature-iteration steps. Each iteration’s state is not only a function of the current feature but also an accumulation of all past features’ effects. In practice, we can formalize this by saying that the state after the $k$-th feature, $h^{(k)}$, contains a logarithmically weighted trace of all previous states $h^{(1)}, h^{(2)}, \ldots, h^{(k-1)}$. This means important patterns introduced early in the sequence leave a residual imprint that continues to influence the trajectory through later iterations.

This intrinsic memory mechanism is crucial for detecting synergistic feature bundles. Combinations of three or more features may jointly separate outcomes strongly, even if each feature alone is only mildly predictive. A traditional greedy algorithm might overlook them, but NetraAI’s memory mechanism locks in partial gains: when multiple features come into play, the system reconverges on the emergent attractor, reinforcing truly predictive, synergistic bundles despite intermediate steps with transient groups caused by noisy features.

By preserving synergistic signals with the built-in memory mechanism, NetraAI prunes the combinatorial feature search — avoiding combinatorial complexity that plagues exhaustive feature selection methods — and focuses on feature combinations that demonstrate durable predictive power. We have previously described how this memory-driven approach amplifies signals that prove consistently predictive over many generations of search, enabling the discovery of patterns that would be infeasible to find via naive search or random exploration (manuscript submitted for review). Moreover, this class of machine learning (ML) has a cross-run knowledge transfer aspect: insights from one dataset or run can prime the system in subsequent runs, further reducing redundancy in exploration.

The configuration space explored by NetraAI exhibits fractal structure, allowing for deep compression of informative patterns across scales.\textsuperscript{8} This is made possible by the underlying dynamical system as it is driven to evolve according to variable sequences, which guide the system along structured paths through high-dimensional feature space. The combination of the dynamical system, pseudo-temporal sequencing, and the long-range memory mechanism yields a multi-scale, self-regularizing learning system: global enough to integrate information across all patients and features while isolating local subgroup structures, balancing exploration of high-order interactions with interpretability and repeatability.

\section{Methodology}

NetraAI is implemented as a two-tier system:
\begin{enumerate}
    \item \textbf{Core Learning Engine:} performs data-driven dynamical contraction mapping, iterative feature-combination learning, and evolutionary model reinforcement.
    \item \textbf{Meta-Evolutionary LLM Strategist:} ingests the learning engine’s outputs, accumulates knowledge across runs, provides biological and medical interpretations, and provides high-level guidance.
\end{enumerate}

The strategist LLM transforms NetraAI from a static analysis engine into a continuously evolving system—one that integrates learning across runs without ever compromising data privacy, as the sponsor’s data remains fully contained within NetraAI. The LLM Strategist contributes memory, contextual reasoning, and high-level strategic guidance, while NetraAI performs the core data-driven discovery. This division of labor—NetraAI as the experimentalist and the LLM Strategist as the theorist—closely mirrors the scientific process. Just as human progress depends on the interplay between empirical observation and theoretical insight, the collaboration between these two AI components creates a closed loop of experimentation and interpretation.

\subsection{Core Learning Engine}

\subsubsection{Data Input}

NetraAI takes as input a clinical trial dataset, where each patient typically includes a set of features (clinical measures, lab values, demographics, -omics data, etc.) and an outcome of interest (e.g., a treatment response or clinical endpoint). Preprocessing is performed to normalize continuous variables, encode categorical variables, and impute missing data as needed.

Optionally, domain knowledge can be supplied in a structured form, such as known risk factors or biomarker groupings. This external knowledge does not directly alter the algorithm’s core learning mechanics but is available to the strategist LLM. In other words, while the core engine is largely data-driven, it is situated in a framework where expert knowledge is readily accessible for guidance.

\subsubsection{Initialization}

Each patient is initially represented by a high-dimensional vector consisting of all available data and embedded into a latent space via the dynamical system. The engine then defines an ordered sequence of features. At each pseudo-time step, each feature-value pair applies a small transformation that drives the system forward in its evolution. In this way, the patient representations co-evolve with the goal of learning feature bundles.

The initial configuration is not important but must be the same for all patients defined by a random initialization seed. This diversity of initialization is important as it produces different evolutionary trajectories through the pseudo-temporal feature-space, sampling multiple ways to stratify the cohort. Unlike methods that rely on a fixed embedding (like PCA or t-SNE) before clustering, NetraAI builds an intrinsic embedding on-the-fly through its dynamics, ensuring that the space is optimally shaped to reflect outcome-relevant structure.\textsuperscript{9}

\subsection{Iterative Contraction Phase}

With the updated sequence in place, NetraAI iteratively applies the contraction mapping $F$ (specifically engineered to allow the encoding and compression of a vast amount of information) to all patient states. After each full cycle through the feature list, the algorithm checks a termination criterion based on how well the resulting model satisfies a loss function. Mathematically, $F$ is designed to satisfy the contraction condition on a metric space $S$ via the Banach Contraction Theorem.\textsuperscript{10} This states that if:

\[
d(F(x), F(y)) < d(x, y) \quad \forall x, y \in S, \, x \neq y
\]

then there exists a unique fixed-point $z$ such that:

\[
\lim_{n \to \infty} F^n(x) = z \quad \forall x \in S
\]

However, we stop the iterations early at the point where the configuration of patient states maximizes an information criterion related to the outcome. Cluster purity-based binary-cross-entropy (BCE) loss is computed on clusters with respect to the outcome label.\textsuperscript{11} At this point, the patients’ positions in the latent space represent a high-resolution stratification of the cohort—those lying close together have been pulled by the same combination of features, likely forming a coherent subgroup.

For clarity, we are given a dataset of $N$ patients (or data samples):

\[
\{(x_i, y_i)\}_{i=1}^N
\]

where each $x_i$ is a representation of patient $i$, and $y_i \in \{0,1\}$ is a binary label (e.g., favorable versus not favorable for a given clinical trial) for patient $i$. This approach partitions the $N$ samples into $K$ distinct clusters in each of the iterated runs. Let $C_k$ denote the set of indices of all samples belonging to cluster $k$, where $k \in \{1, 2, \ldots, K\}$, and:

\[
\bigcup_{k=1}^{K} C_k = \{1, 2, \ldots, N\}, \quad C_j \cap C_k = \emptyset \quad \text{for } j \neq k
\]

Define the purity of cluster $k$ with respect to the positive class ($y=1$) as:

\[
p_k = \frac{\sum_{i \in C_k} \mathbb{1}[y_i = 1]}{|C_k|}
\]

where $\mathbb{1}[\cdot]$ is the indicator function and $|C_k|$ is the total number of points in cluster $k$. The value $p_k$ thus represents the fraction of samples in cluster $k$ that have the label $y = 1$. Each patient $x_i$ in cluster $k$ will have its predicted probability of being in the positive class given by:

\[
p_i = p_k, \quad \text{for each } i \in C_k
\]

Hence, all samples in the same cluster inherit the same predicted probability $p_k$. Given the cluster-induced probabilities $\{p_i\}$, the BCE loss over the entire dataset is:

\[
L_{\text{BCE}} = -\sum_{i=1}^N \left[y_i \ln(p_i) + (1 - y_i) \ln(1 - p_i)\right]
\]

Since $p_i = p_k$ for $i \in C_k$, we can equivalently write:

\[
L_{\text{BCE}} = -\sum_{k=1}^K \sum_{i \in C_k} \left[y_i \ln(p_k) + (1 - y_i) \ln(1 - p_k)\right]
\]

This function is used to evaluate models, and the elimination of poor models through an evolutionary process is the means by which combinatorial feature selection is performed. The long-range memory mechanism is what allows the computational complexity of this evolutionary process to be greatly reduced.

\subsection{Extraction of Patient Categories (Personas)}

Once this intermediate convergence is reached, the technology examines the resulting embeddings of the patients that survive the process. Since the contraction mapping inherently groups patients according to synergistic combinations of variables, proximity in the latent space implies shared key combinations of outcome-driving features. Next, an analysis on the embedding space from multiple geometric perspectives and resolutions identifies a set of $N$ variables that consistently differentiate patient classes across these views. This initial process massively reduces and prioritizes the search space, focusing subsequent analysis on the most informative dimensions for subgroup discovery.

NetraAI then exhaustively evaluates combinations of 2–4 variables and their corresponding value intervals based on their ability to define a subpopulation of patients that exhibit a favorable clinical outcome for the therapeutic intervention under strict statistical constraints (effect size, p-value) to avoid overfitting. Surviving combinations become \textbf{Personas} — compact, interpretable clinical signatures tied directly to therapeutic benefit and suitable for downstream actions such as inclusion/exclusion criteria or biomarker hypotheses.

\subsection{Evaluation and Selection}

The quality of the candidate persona partition is evaluated based on clinical and statistical criteria:

\begin{enumerate}
    \item \textbf{Predictive Utility:} Does separating patients according to these Personas significantly improve outcome prediction compared to the full cohort?
    \item \textbf{Statistical Significance:} Are the outcome differences between the Personas likely to be real (e.g., p-values for outcome contrasts between groups)?
    \item \textbf{Simplicity:} Are the Personas defined by a minimal, non-redundant set of features?
\end{enumerate}

Personas with large effect sizes and robust statistics are preferred. These are derived by using the Persona to select the corresponding patient population, and evaluating how well this population separates control versus drug.

\subsection{Evolutionary Update (Crossover of Solutions)}

Multiple runs of the contraction-based learning process with varied initial conditions and hyperparameters generate an evolving population of candidate solutions, retaining those with the strongest outcome-based signals. An internal form of crossover to further refine the hypothesis space is performed—high-performing feature bundles or subgroup configurations across runs are merged to propose new candidates. For example, if one run identifies \{Feature A high, Feature B low\} as defining a strong responder group, and another finds \{Feature C present, Feature D low\}, NetraAI may test a solution that preserves both subgroups, if they are distinct, or explore hybrid combinations.

This evolutionary process is recursive and self-directed—fresh runs provide built-in variation without requiring explicit mutation steps. The system refines its internal search space, high-value feature combinations persist and propagate, while low-signal structures naturally fade. This results in a distilled set of Personas that consistently explain outcome variation and survive internal replication.

\subsection{Output and Interpretation}

The final output of the Core Learning Engine is a small set of Personas, each with defining variables and ranges, and cohort size and outcome statistics (e.g., response rates, p-values).

For example, NetraAI might identify the following three Personas in a given trial:

\begin{itemize}
    \item \textbf{Persona 1:} $\sim$15\% of patients — characterized by Feature A $>$ 1.2 and Feature B $<$ 0.5, with a median age of 60. This subgroup had an 80\% treatment response rate versus 50\% in the overall cohort ($p = 0.005$).
    \item \textbf{Persona 2:} $\sim$10\% of patients — characterized by presence of Biomarker X and predominantly female gender. This subgroup had only a 30\% response rate (non-responders), significantly below baseline ($p = 0.03$).
    \item \textbf{Persona 3:} $\sim$75\% of patients — all remaining patients not fitting the above profiles, with no distinguishing feature pattern and a response rate around the baseline 50\%.
\end{itemize}

These personas are immediately actionable:
\begin{itemize}
    \item \textbf{Persona 1:} a subset of patients who benefit substantially from the treatment. This group could be targeted in future trials or clinical practice for enrichment.
    \item \textbf{Persona 2:} a group that does poorly and perhaps requires an alternative therapy or closer monitoring.
    \item \textbf{Persona 3:} represents the general population for whom the treatment has average effect.
\end{itemize}

Importantly, NetraAI is designed not to force every patient into an artificial subgroup. An “uncategorized” category, like Persona 3, is used for portions of the cohort that do not form any high-confidence pattern, often occurring in noisy biological data. Traditional neural networks will try to fit all data and can mask heterogeneous effects by averaging, while NetraAI explicitly isolates only the strong signals and lets the rest remain as background. This reflects a core philosophy—it is better to acknowledge uncertainty by leaving some patients unclassified than to overfit and invent spurious clusters. What matters is not perfect individual prediction, but the discovery of a simple, actionable characterization whose application enriches the trial population toward greater overall success. Even if imperfect, this enrichment effect can shift the outcome distribution enough to influence whether a trial achieves its primary endpoints.

\section{Meta-Evolutionary LLM Strategist}

While the above steps describe the inner workings of NetraAI’s Core Learning Engine independently discovering statistically and clinically meaningful subpopulations, its utility is amplified through interaction with an LLM serving as a strategist. Modern LLMs trained on vast biomedical literature and domain knowledge excel at knowledge integration and hypothesis generation; however, they are not designed to find latent structure in raw clinical datasets—this is NetraAI’s role.\textsuperscript{12,13} Thus, the LLM is assigned a complementary role: coordinator and interpreter across multiple NetraAI runs—retaining memory of the NetraAI run outputs and accumulating a knowledge base of what patterns have emerged, guiding future analyses:

\begin{enumerate}
    \item \textbf{Strategic Seeding of Variable Sequences:} The LLM strategist may prioritize variables that frequently appear in high-performing Personas to accelerate convergence and refine the exploration of high-value regions in the configuration space. This is done to test for stable reproducibility to deepen signal amplification, not to bias results. Using external domain knowledge, the LLM can also prioritize variables known to correspond to clinically recognized subtypes to guide NetraAI toward medically relevant distinctions without constraining its discovery process.
    
    \item \textbf{Domain Expertise:} The LLM strategist can draw on medical knowledge to steer NetraAI in plausible directions. For example, the LLM might recognize a link to a known mechanism from the literature and suggest exploring specific variables in subsequent runs.
    
    \item \textbf{Robustness Assessment:} By tracking recurring patterns across runs, the LLM strategist can flag stable subgroups versus one-off artifact. This adds a layer of validation—a Persona the LLM strategist has “seen” multiple times in slightly different forms is more likely to be genuine.
    
    \item \textbf{Global Coordination:} The LLM strategist can perform a meta-analysis-like function, sharing insights across parallel NetraAI instances on different slices of data for large trials with many possible sub-cohorts. For example, a pattern found in one sub-cohort could be suggested to another.
\end{enumerate}

After a NetraAI analysis, we have a set of Personas and their performance metrics, and the LLM Strategist can generate a narrative report. For example, “Future clinical trials might benefit from assessing the relative severity of these symptom domains as potential predictors of treatment response. Alternative mechanisms targeting both sleep and detachment symptoms might be more effective for this subgroup.” Such a summary not only provides human users with context but can also feed back into the next computational cycle.

\subsection{Self-Improving System}

NetraAI, a dynamical learner coupled with the LLM Strategist, a reflective language model, yields a self-improving system that evolves over time, unlike static models that are trained once and then fixed. After each NetraAI run, the LLM Strategist updates its contextual understanding of what variables and combinations appear important. While the LLM’s internal weights might not change (unless explicitly fine-tuned), it effectively learns through conversation with NetraAI’s results—much like a human analyst learning from new data insights. Over multiple cycles, the LLM Strategist develops an increasingly rich “theory” of the dataset and similar datasets and may start to anticipate what kinds of Personas are plausible or interesting even before NetraAI runs. 

This does not mean it controls NetraAI or biases it to confirm prior beliefs unjustifiably; rather, it acts as a guide to help NetraAI navigate the search space more efficiently and intelligently than random restarts, and as an interpreter to ensure that findings are placed in the context of existing scientific knowledge. In essence, the LLM provides a form of theory-of-mind for the NetraAI process, as it learns from the NetraAI insights and provides guidance in a way that a standalone algorithm cannot.

\subsection{Transparency and Auditability}

All LLM Strategist suggestions are logged for transparency and auditability, aligning with verification, validation, and uncertainty quantification (VVUQ) principles for reliable AI. A record of why certain analyses were pursued is available for regulators or researchers, ensuring that the meta-level decisions can be scrutinized just like the model outputs. For example, if the LLM Strategist prompted an analysis focusing on “patients with feature X in a certain range,” that rationale is recorded (“due to recurring pattern of feature X in previous findings, or relevant literature Y”).

\subsection{Fine-Tuning}

Although the LLM is not involved in the core learning process, it accumulates contextual knowledge across runs, allowing it to track recurring patterns, refine hypotheses, and help guide future analyses. Over time, the LLM Strategist becomes exposed to mechanistic insights it could not generate on its own, allowing it to mature as an interpreter of NetraAI’s discoveries. In this way, the system gains a reflective capacity: it not only learns patterns in the data but gradually improves its own understanding of what types of explanations hold up across time, datasets, and domains. This is a form of soft fine-tuning: the LLM is updating its understanding of clinical trials through the lifting of insights that NetraAI can deliver, even if its internal weights remain the same.

Future work will explore fine-tuning the LLM Strategist using a structured tokenization of NetraAI-derived insights. Each Persona can be encoded as a compact, high-signal token sequence that captures variable bundles, threshold conditions, outcome alignment, replication history, and trial context, for example:

\begin{quote}
\ttfamily
[DISEASE=Schizophrenia] [VAR\_1=COWAT>12] [VAR\_2=Curiosity=High] [OUTCOME=Responder] [P\_VALUE=0.004]
\end{quote}

By collecting tens of thousands of tokens across diverse trials and indications, we can build a rich, structured corpus to fine-tune the LLM Strategist’s weights.

This allows the LLM to internalize the logic of outcome-driven subpopulation discovery, enhancing its ability to interpret, explain, and even anticipate Persona structures in future runs. Over time, this transforms the LLM Strategist from a passive interpreter into an adaptive strategist, capable of reasoning with empirical knowledge grounded not just in literature, but in validated, data-derived insight—further closing the loop between empirical discovery and theoretical insight, providing a stark advantage over current LLM models being offered to help clinical trialists due to the superior and refined resolution of insights captured through NetraAI.

\section{Experimental Evaluation with Case Studies}

We evaluated NetraAI on three real-world clinical trial datasets, comparing the performance of its Persona-driven insights against (a) standard ML models and (b) direct LLM analysis, to demonstrate NetraAI’s superior ability to uncover meaningful subpopulations and improve predictive modeling in complex, heterogeneous clinical data. For fairness, all methods were scored on predictive performance rather than purely Persona discovery. Across each case, NetraAI’s small, interpretable subpopulations and feature learning significantly boosted model robustness.

A key component in this success is NetraAI’s selective classification strategy—deliberately leaving a fraction of patients “No Call” to focus on reliably predictable subgroups. In this section we outline the impact of NetraAI augmented by the LLM Strategist.

\subsection{Case Study 1: CATIE Schizophrenia Trial}

\subsubsection*{Dataset and Outcome}

Using the CATIE schizophrenia trial ($n \approx 1600$) evaluating the effectiveness of several antipsychotics, we focused on two of the treatment arms (perphenazine and olanzapine) to analyze differential outcomes.\textsuperscript{14} Each patient had 292 baseline features spanning multiple domains: psychiatric symptom severity (e.g., PANSS, CGI scores), functional status (questionnaires like SF-12, QLS), side effect scales (AIMS, SAS for movement side effects; metabolic panels for weight gain, etc.), neurocognitive assessments, and basic demographics and labs.

We defined the outcome as \textit{time to all-cause treatment failure}, operationalized as whether a patient discontinued their assigned medication and switched to a different one (within an 18-month period). This outcome captures both efficacy and tolerability—whether the treatment “worked” for the patient or not.

\subsubsection*{Analysis Objectives and Methods}

We aimed to determine whether small, explainable subpopulations within the schizophrenia cohort could be distinguished using NetraAI’s feature discovery and relabeling. Specifically, we evaluated:

\begin{enumerate}
    \item \textbf{Traditional ML Performance:} Using standard classification models (random forest, naïve Bayes, gradient boosting, and deep neural networks) on the full feature set and with feature selection via traditional methods (LASSO regularization, filter methods like ANOVA F-statistics, and SHAP values) to identify the top 20 features.\textsuperscript{15}
    
    \item \textbf{NetraAI Pipeline Improvement Gain:} Using NetraAI’s Persona discovery to identify key variables and applying NetraAI’s relabeling and feature selection across multiple ML classifiers.
    
    \item \textbf{LLM Analysis:} Several LLMs, including DeepSeek, were evaluated for whether purely language-based AI could interpret or classify any subpopulations.
\end{enumerate}

A reduced subset ($n = 52$) was used across the three analyses for fairness due to context window and dimensionality limitations of the LLMs. The goal was to distinguish patients that were perphenazine completed or olanzapine failed (PCOF; $n = 21$) from those that were perphenazine failed or olanzapine completed (PFOC; $n = 31$), a separation not explicitly encoded in the original labels.

\subsubsection{Results}

\subsubsection*{Traditional ML Performance}

Using the full feature set, conventional ML models achieved only modest accuracy (around 55--65\% on predicting treatment failure, with AUC $\approx$ 0.5--0.68). We attribute these near-chance statistics to the high dimensionality and heterogeneity—the signal was diluted across many variables.

Feature selection via traditional methods was applied to reduce dimensionality (e.g., top 20 features), resulting in improved models with the logistic regression model achieving $\sim$65\% accuracy and the XGBoost model achieving $\sim$66\% accuracy (AUC of $\sim$0.68).

These poor results underscore the difficulty of the task—no straightforward pattern jumps out with conventional analysis, and interpretability remains low as the models are essentially black boxes on a subset of features.

\subsubsection*{NetraAI Pipeline Improvement Gain}

Using 5-fold stratified cross-validation holding out 20\% in each fold, NetraAI produced the following relabeling and subpopulation structure:

\begin{itemize}
    \item \textbf{Subpopulation A:} $n = 10$ (7 PCOF, 3 PFOC)
    \item \textbf{Subpopulation B:} $n = 11$ (1 PCOF, 10 PFOC)
    \item \textbf{Subpopulation C:} Unclassified (“No Call”); $n = 31$
\end{itemize}

NetraAI identified a high-performing feature bundle of four variables that optimally separated the two relabeled subgroups:

\begin{itemize}
    \item \textbf{COWAT:} Number of Correct Words Generated (verbal fluency)
    \item \textbf{QOL:} Self-Rated Curiosity
    \item \textbf{QOL:} Moderate Vocational Activity
    \item \textbf{QOL:} Extremely Restrictive Living Environment
\end{itemize}

These variables were used to construct simple, interpretable models that outperformed baseline approaches, despite the small dataset (Table~\ref{tab:model-performance}).

\begin{table*}[t]
\centering
\caption{Model performance across classifiers using baseline features, NetraAI-guided feature selection, and NetraAI-guided feature selection with subpopulation relabeling}
\vspace{0.5em}
\label{tab:model-performance}
\begin{tabularx}{\textwidth}{l|XXX|XXX}
\toprule
\textbf{Model} & \multicolumn{3}{c|}{\textbf{AUC}} & \multicolumn{3}{c}{\textbf{Accuracy}} \\
              & \textbf{Baseline} & \textbf{NetraAI 4 Vars} & \textbf{NetraAI 4 Vars + Subpop} 
              & \textbf{Baseline} & \textbf{NetraAI 4 Vars} & \textbf{NetraAI 4 Vars + Subpop} \\
\midrule
Random Forest       & 0.242 & 0.704 & 1.000 & 0.65 & 0.70 & 0.95 \\
Naïve Bayes         & 0.456 & 0.694 & 1.000 & 0.55 & 0.68 & 1.00 \\
Gradient Boosting   & 0.297 & 0.574 & 1.000 & 0.40 & 0.56 & 1.00 \\
Deep Neural Network & 0.516 & 0.594 & 0.990 & 0.60 & 0.62 & 0.95 \\
\bottomrule
\end{tabularx}
\end{table*}

\subsubsection*{LLM Analysis}

LLMs lacked the capacity to independently extract or classify meaningful structures from this dataset, reinforcing NetraAI’s value as a mathematically-grounded discovery engine for high-dimensional clinical data tasks. After carefully working with DeepSeek to explain that the goal was to find patients that would perform better on one drug over the other, that any ML technique could be used, and even recommending methods, it was only able to find one subpopulation of one patient.

\subsubsection*{Summary}

NetraAI identified high-effect-size, explainable subpopulations in a clinical trial with limited sample sizes that transformed weak baseline models into near-perfect classifiers, showcasing its strength in low-signal, high-dimensional psychiatric data. By combining compact variable sets with selective relabeling, NetraAI is a powerful tool for refining trial insights, particularly in domains where patient heterogeneity and sparse labels challenge traditional ML approaches.

\subsection{Case Study 2: CAN-BIND Major Depressive Disorder (MDD) Trial}

\subsubsection*{Dataset and Outcome}

Using the CAN-BIND major depressive disorder (MDD) trial, we used the exploratory escitalopram arm ($n = 172$) to analyze preferential treatment response.\textsuperscript{16} Each patient had over 362 clinical and behavioral baseline features including assessments such as CGI, SEXFX, DARS, SHAPS, MINI, MADRS, YMRS, QIDS, PSQI, and SPAQ. We defined the outcome as a $\geq 50\%$ reduction in MADRS total score from baseline over 8 weeks. This outcome captures efficacy of escitalopram for improving symptoms for the patient or not.

\subsubsection*{Analysis Objectives and Methods}

We aimed to determine whether small, explainable escitalopram responsive subpopulations within the MDD cohort could be distinguished using NetraAI’s feature discovery and relabeling. Specifically, we evaluated:

\begin{enumerate}
    \item \textbf{Traditional ML Performance:} Using standard classification models (logistic regression, XGBoost, random forest, support vector machine, deep neural networks) on the full feature set and with feature selection via traditional methods (LASSO regularization, filter methods like ANOVA F-statistics, and SHAP values) to identify the top 20 features.\textsuperscript{15}
    \item \textbf{NetraAI Pipeline Improvement Gain:} Using NetraAI’s Persona discovery to identify key variables and applying NetraAI’s relabeling and feature selection across multiple ML classifiers.
\end{enumerate}

\subsubsection*{Results}

\paragraph{Traditional ML Performance.} Using the full feature set ($n = 173$; 90 non-responders, 83 responders), conventional ML models achieved limited performance, with classification accuracies ranging from 55.8\% to 65.7\%, and AUC values between 0.49 to 0.68. The best model was XGBoost with ANOVA F-statistic using 20 features (Table~\ref{tab:canbind-xgb}).

\begin{table*}[t]
\centering
\caption{Accuracy of best traditional ML – XGBoost (ANOVA F-statistic with 20 features)}
\vspace{0.5em}
\label{tab:canbind-xgb}
\begin{tabularx}{\textwidth}{lX}
\toprule
\textbf{Metric} & \textbf{Value} \\
\midrule
Accuracy (\%) & 65.71 \\
Accuracy 5-Fold CV (\% $\pm$ SD) & 61.43 $\pm$ 6.09 \\
Accuracy 10-Fold CV (\% $\pm$ SD) & 56.76 $\pm$ 10.36 \\
Accuracy LOO CV (\% $\pm$ SD) & 54.39 $\pm$ 4.98 \\
\bottomrule
\end{tabularx}
\end{table*}

These models struggled to generalize due to the high dimensionality and heterogeneity of the dataset, resulting in suboptimal sensitivity and specificity.

\paragraph{NetraAI Pipeline Improvement Gain.} Using 5-fold stratified cross-validation holding out 20\% in each fold, NetraAI produced the following relabeling structure:

\begin{itemize}
    \item \textbf{Subpopulation A:} $n = 25$ (Non-Responders)
    \item \textbf{Subpopulation B:} $n = 23$ (Responders)
    \item \textbf{Subpopulation C:} $n = 115$ (Unclassified/No Call)
\end{itemize}

NetraAI identified a high-performing feature bundle of eight anhedonia–neurovegetative variables that optimally separated the two relabeled subgroups:

\begin{itemize}
    \item \textbf{SHAPS:} I would enjoy my favorite meal (anhedonia)
    \item \textbf{QLESQ:} Physical health (physical health satisfaction)
    \item \textbf{QLESQ:} Household activities (daily functioning)
    \item \textbf{DARSD:} Desire to spend time with family (social anhedonia)
    \item \textbf{DARSD:} Desire to spend time with friends (social anhedonia)
    \item \textbf{DARSD:} Desire to meet new people (social motivation)
    \item \textbf{MADRS:} Reduced appetite (neurovegetative)
    \item \textbf{MADRS:} Pessimistic thoughts (cognition)
\end{itemize}

Using the NetraAI findings improved the performance of all evaluated ML models, with the most notable improvement in SVM (Table~\ref{tab:canbind-improvement}). In addition to accuracy:

\begin{itemize}
    \item Sensitivity increased by $\sim 31\%$
    \item Specificity improved by $\sim 51\%$
    \item F1-score increased by $\sim 30\%$
    \item AUC rose by $\sim 39\%$ with values approaching 0.99
\end{itemize}

\begin{table*}[t]
\centering
\caption{Traditional model performance across classifiers before and after NetraAI insights to capture improvement gain}
\vspace{0.5em}
\label{tab:canbind-improvement}
\begin{tabularx}{\textwidth}{lXXX}
\toprule
\textbf{Model} & \textbf{Accuracy Before NetraAI (\%)} & \textbf{Accuracy After NetraAI (\%)} & \textbf{Improvement (\%)} \\
\midrule
Logistic Regression & 54.29 & 77.14 & +22.85 \\
XGBoost             & 65.71 & 91.43 & +25.72 \\
Random Forest       & 62.86 & 82.86 & +20.00 \\
SVM                 & 60.00 & 100.00 & +40.00 \\
Deep Neural Network & 60.00 & 77.14 & +17.14 \\
\bottomrule
\end{tabularx}
\end{table*}

Notably, NetraAI’s “No Call” class accounted for $\sim$70\% of the patient population, with the remaining 30\% forming stable, high-confidence subgroups. Attempts to utilize LLMs directly, even with guidance, were unable to produce meaningful subpopulations.

\subsubsection*{Summary}

NetraAI isolated high-effect-size, explainable subpopulations while introducing an “unknown class” to leave ambiguous cases unclassified to better handle data uncertainty. This reduction to compact feature sets mitigated overfitting and improved generalizability across models, trading complete coverage for reliability in severe clinical depression data.

This deliberate trade-off reflects NetraAI’s core design philosophy: not all patients can be reliably modeled in small, noisy clinical trial datasets. The goal is not total prediction, but the extraction of meaningful, replicable subpopulation insights that can be used to inform enrichment strategies in future trials.

\subsection{Case Study 3: COMPASS Pancreatic Ductal Adenocarcinoma (PDAC) Trial}

\subsubsection*{Dataset and Outcome}

Using the COMPASS pancreatic ductal adenocarcinoma (PDAC) trial ($n = 87$) to identify predictive mutational and transcriptional features for better treatment selection, we focused on the Gemcitabine plus nab-paclitaxel (GnP) and FOLFIRINOX (FFX) first-line chemotherapy arms.\textsuperscript{17} Each patient had approximately 25{,}000 single nucleotide variants (SNVs) and transcriptomic features.

We defined outcome as tumor response evaluated using standard criteria:

\begin{itemize}
    \item \textbf{Stable Disease (SD):} Tumor growth $\leq 20\%$ or reduction $< 30\%$
    \item \textbf{Partial Response (PR):} Tumor reduction between 30\% and 100\%
    \item \textbf{Complete Response (CR):} Full disappearance of the tumor
\end{itemize}

\subsubsection*{Analysis Objectives and Methods}

We aimed to determine whether small, explainable responsive subpopulations to FFX or GnP within the PDAC cohort could be distinguished using NetraAI’s feature discovery and relabeling. Specifically, we evaluated:

\begin{enumerate}
    \item \textbf{Traditional ML Performance:} Using standard classification models (gradient boosting, SVM, deep neural network, naïve Bayes) on the full feature set and with univariate feature selection via ANOVA F-test to select the top 15 SNVs to train traditional classifiers to predict treatment-specific benefit (GnP versus FFX).\textsuperscript{15}
    \item \textbf{NetraAI Pipeline Improvement Gain:} Using NetraAI’s Persona discovery to identify key variables and applying NetraAI’s relabeling and feature selection across multiple ML classifiers.
\end{enumerate}

\subsubsection*{Results}

\paragraph{Traditional ML Performance.} Using the full feature set, conventional models achieved only modest performance. Gradient boosting evaluated using nested 5-fold cross-validation was the best performing model (Table~\ref{tab:compass-gb}), highlighting the difficulty of predictive modeling in high-dimensional, small-cohort oncology datasets.

\begin{table*}[t]
\centering
\caption{Best performing traditional ML – Gradient Boosting with nested 5-fold CV}
\vspace{0.5em}
\label{tab:compass-gb}
\begin{tabularx}{\textwidth}{lXXXXX}
\toprule
\textbf{Metric} & \textbf{Accuracy (\%)} & \textbf{F1 Score} & \textbf{Sensitivity} & \textbf{Specificity} & \textbf{AUC} \\
\midrule
Gradient Boosting & 53.0 & 0.562 & 61.0\% & 45.2\% & 0.569 \\
\bottomrule
\end{tabularx}
\end{table*}

Models trained on the raw high-dimensional SNV space demonstrated limited separability of treatment preference, poor generalization, and failed to align directionally with actual tumor shrinkage.

\paragraph{NetraAI Pipeline Improvement Gain.} NetraAI identified a clinically meaningful subpopulation ($n = 23$; 13 FFX, 10 GnP) that perfectly differentiated treatment response between patients who benefited from FFX and those who responded to GnP, characterized by three SNVs:

\begin{itemize}
    \item \textbf{ZYFVE20}
    \item \textbf{LOC389043}
    \item \textbf{MIR30D}
\end{itemize}

Using these three features and NetraAI-assigned labels:

\begin{itemize}
    \item FFX Responsive
    \item GnP Responsive
    \item No-Call
\end{itemize}

Classification models were retrained, and all models achieved perfect generalization performance:

\begin{itemize}
    \item \textbf{AUC:} 1.00
    \item \textbf{Accuracy:} 1.00
    \item \textbf{Separability:} Complete for both binary (0 vs 1) and multiclass (0, 1, 2) tasks
\end{itemize}

As a result, NetraAI successfully isolated the only generalizable subpopulation in the dataset, discovering a minimal and interpretable SNV signature that robustly predicts treatment-specific outcomes.

To assess whether the NetraAI-identified signatures correspond to true clinical benefit, we computed the \textit{C-for-benefit} metric. A gradient boosting classifier was trained using the three NetraAI-selected SNVs, focusing only on patients with confident treatment preference predictions (FFX-responsive vs GnP-responsive). For each matched pair (based on Euclidean distance in the SNV feature space), we compared the predicted benefit difference to the actual difference in tumor shrinkage. The resulting \textbf{C-for-benefit score was 0.923}, indicating that 92.3\% of paired comparisons reflected the real-world treatment benefit.

Attempts at guiding LLMs were not successful. However, the \textit{C-for-benefit} result was achievable only if DeepSeek (or other LLMs) were explicitly given the results discovered by NetraAI.

\subsubsection*{Summary}

NetraAI identified an interpretable 3-SNV biomarker signature that perfectly separated response to GnP and FFX, validating its power as a reliable tool for guiding treatment decisions in high-dimensional oncology datasets.

\section{Selective Classification Trade-Off and Persona Example}

NetraAI’s superior subgroup discovery hinges on a deliberate trade-off: introducing an “unknown” class for a portion of the patient population that the model cannot predict with high confidence. This is not a failure, but rather a recognition of the inherent limitations of small, noisy clinical trial datasets. Trials often involve small samples and incomplete representation of disease heterogeneity; expecting any model to generalize across the full biological distribution risks overfitting or spurious clusters.

Instead, NetraAI relabels the population in a way that focuses predictive attention on subgroups that exhibit consistent, high-effect-size patterns, leaving patients without reproducible signals “uncalled” to prioritize reproducibility and interpretability over full coverage.

By narrowing its scope to prioritize the most reliable and reproducible cohorts—with a clear trade-off that not all patients will be classified—NetraAI yields clinically interpretable groups with strong outcome associations. In practice, NetraAI may identify only one—or several—explainable subpopulations, each accompanied by strong statistical validity, while the remainder remain unclassified.

The practical implication of this selective learning strategy is profound: NetraAI is not built to explain everything, but rather to learn enough about the clinical trial to inform future enrichment. By isolating and explaining the subgroup(s) most likely to benefit from the investigational drug, NetraAI enables actionable insights and robust hypothesis generation even when global prediction is impossible. This allows sponsors to refine inclusion criteria or develop targeted biomarker strategies that increase the probability of success in follow-up trials.

This ability to precisely identify advantage-aligned subpopulations in the presence of small, noisy datasets distinguishes NetraAI from conventional ML approaches that strive for uniform prediction but often obscure the signals that matter most.

\subsection{NetraAI Persona with LLM Augmentation Example}

To illustrate NetraAI’s explanatory power and LLM augmentation, consider the following representative Persona discovered in a prior client engagement. While the dataset and sponsor must remain confidential, the Persona represents the type of robust, high-signal subpopulations NetraAI consistently discovers in clinical trial data:

\paragraph{Persona A: High-Functioning, Inflammation-Sensitive Responders}

\textbf{Defining Variables:}
\begin{itemize}
    \item Depression Severity Score: Moderate baseline severity (10–14)
    \item Motivation Subscale: Above-average motivation ($>4.5$)
    \item C-Reactive Protein (CRP): Elevated systemic inflammation ($>3.0$ mg/L)
\end{itemize}

\textbf{Persona Characteristics:}
\begin{itemize}
    \item Size: 13–15\% of the trial cohort
    \item Response Rate: 85–90\% versus 55\% overall ($p < 0.01$)
\end{itemize}

NetraAI converged on this Persona through multiple runs, leveraging its pseudo-temporal embedding and long-range memory to isolate stable, outcome-relevant attractors. Although inflammation is often associated with treatment resistance, preserved motivational capacity emerged as a distinguishing and predictive trait, delineating a subgroup with exceptional treatment benefit.

\paragraph{LLM Strategist Commentary}

\begin{quote}
\emph{
“This persona captures a clinically meaningful intersection between systemic inflammation and preserved engagement. Although inflammation is frequently linked with SSRI non-response, elevated motivation scores may indicate compensatory neural resilience. The strong effect size observed here suggests that motivation can mediate treatment outcomes in high-CRP patients. This warrants further investigation, especially in trials assessing the synergistic impact of anti-inflammatory adjuncts or motivational enhancement strategies.”
}
\end{quote}

This example underscores how NetraAI’s selective classification, combined with LLM-contextualization, generates concise, actionable hypotheses for trial enrichment while transparently acknowledging areas of uncertainty.

\section{Discussion}

\subsection{Complementary Strengths of AI Species}

The interaction of NetraAI with the LLM Strategist reflects a broader design principle—occupying distinct but complementary roles. LLMs excel at integrating knowledge, hypothesis generation, and interpreting unstructured information. In contrast, NetraAI delivers rigorous, interpretable subpopulation discovery from structured clinical trial data via its contraction dynamics and evolutionary feature search.

By coupling NetraAI’s Persona embeddings (compact, concept-level subgroups) with the LLM Strategist’s contextual reasoning over abstract tokens and clinical concepts, we mirror the human scientific process: the LLM acts as a theorist, contextualizing and proposing directions, while NetraAI acts as an experimentalist, providing testable structure from data. Rather than competing, these components form a co-evolving, symbiotic loop where verified signal informs reasoning, and reasoning guides further search—charting a path toward the next generation of agentic, interpretable AI for scientific domains.

\subsection{Interpretability and Trust in Clinical AI}

Clinical decisions carry high stakes, and black-box models face understandable skepticism from practitioners and regulators that demand explainable AI in medicine. NetraAI addresses this by producing immediately interpretable and actionable hypotheses. Each Persona is a hypothesis that can be evaluated and tested in the real world.

For example, a NetraAI output might be summarized as:

\begin{quote}
\emph{“Patients with biomarker X above 10 and a high fatigue score benefited markedly from the drug, whereas patients with low X and high anxiety did not.”}
\end{quote}

Such statements are intelligible to clinicians and can be followed up (e.g., check biomarker X levels in new patients, consider interventions for high anxiety patients, etc.). Each Persona is accompanied by statistics (p-values, effect sizes, and confidence intervals), matching the statistical rigor of clinical research.

Moreover, NetraAI’s internal validation (required replication across bootstrap resamples or splits) and cross-run strategist memory ensure that strong Personas reappear under varied conditions—mirroring the scientific practice of reproducing findings across subsets. This ensures that identified subgroups are not one-off flukes and confirms replicability. By the time we present a Persona externally, we have considerable evidence that it’s real.

This built-in interpretability goes far beyond typical post-hoc explanations of black-box models, fostering the trust necessary for regulatory acceptance. In heavily regulated environments like drug development, traceability and replicability are invaluable—every decision and finding can be audited and is supported by explicit evidence, making the AI’s conclusions more palatable for review and adoption.

\subsection{Comparison with Alternative AI Paradigms}

Here, we contrast NetraAI with other prominent AI approaches to highlight its unique strengths in the context of clinical data. While reinforcement learning, symbolic systems, tree-based models, and classical clustering each offer valuable techniques, they often lack NetraAI’s combination of outcome-awareness, interpretability, and robust handling of heterogeneity.

\paragraph{Reinforcement Learning.} Reinforcement learning is powerful in interactive environments but is ill-suited to static trial datasets. Clinical trials do not offer sequential decision-making environments typical of reinforcement learning domains, with even adaptive trials involving constrained and predefined decision spaces. NetraAI’s model is closer to supervised and unsupervised learning, with interpretability as a core feature. Its evolutionary search borrows the explore–exploit trade-off without policy learning. In principle, NetraAI could inform reinforcement learning systems by identifying meaningful subpopulations that map to distinct policy regimes.

\paragraph{Symbolic AI and Knowledge Graphs.} Traditional symbolic systems encode logical rules and inference paths but struggle with uncertainty and scale. In contrast, NetraAI derives empirical symbols (Personas) directly from data, with each Persona becoming a concept grounded in observed outcomes. NetraAI’s outputs can be used to construct knowledge graphs, connecting variables, outcomes, and Personas—hybridizing data-driven discovery and symbolic representation, and aligning NetraAI with neuro-symbolic AI approaches.

\paragraph{Tree-Based Machine Learning.} Decision trees segment data by greedy, single-feature splits, resulting in numerous fragile subgroups. NetraAI takes a more holistic path, identifying multi-feature attractors and allowing an “unknown” class to avoid overfitting. Unlike decision trees or random forests, which may lack clear interpretability at the ensemble level, NetraAI prunes its output through evolutionary learning to result in a small, robust set of interpretable Personas.

\paragraph{Clustering and Pattern Mining.} Classical clustering algorithms do not account for outcomes, and subgroup discovery methods often rely on brute-force or greedy heuristics. NetraAI integrates contraction-based grouping with outcome-aware fitness, achieving a balance between unsupervised structure discovery and supervised relevance in a computationally tractable iterative process.

\section{Uncertainty Quantification and Validation}

NetraAI integrates statistical rigor via bootstrap resampling, confidence intervals, p-values, and effect size. Personas that replicate across bootstrap samples demonstrate stability, while unstable ones are discarded. The LLM Strategist’s logged recommendations allow for transparent auditing of the learning process, further supporting verifiability. Together, these practices align with verification, validation, and uncertainty quantification (VVUQ) principles—critical for any AI deployed in regulated clinical settings.

\section{Limitations and Challenges}

We acknowledge current limitations and challenges inherent to NetraAI’s design and outline areas for future enhancement—including support for diverse outcome types, managing LLM variability, and addressing computational scaling—to guide ongoing development and responsible deployment.

\begin{itemize}
    \item \textbf{Outcome Types:} Currently, NetraAI focuses on categorical binary outcomes. Extensions to continuous measures or survival endpoints (e.g., variance-explained fitness metrics, log-rank tests) are underway.
    
    \item \textbf{LLM Variability:} Reliance on an LLM Strategist introduces variability—while guided by domain prompts, the LLM can occasionally propose implausible directions. This is mitigated with constrained prompting and human oversight.
    
    \item \textbf{Computational Complexity:} While NetraAI is designed for hypothesis generation and not confirmation, rigorous validation remains essential. For high-dimensional inputs, computational complexity grows quickly—pairwise distances scale quadratically with sample size. NetraAI’s long-range memory mechanism for combinatorial feature learning, dimensionality reduction, and feature sparsification help contain this growth.
\end{itemize}

\section{Implications for Clinical Trials and Personalized Medicine}

NetraAI’s capacity to find and validate subpopulations can improve clinical trial design and analysis. In trial analysis, it can identify responder subgroups that can drive adaptive trial enrichment—such as focusing enrollment on promising subgroups mid-trial. Additionally, NetraAI aids in post-hoc analyses for regulatory submissions, where demonstrating that a drug works particularly well in a specific defined subgroup (and not worse in others) can sometimes support approval or further investigation.

In personalized medicine, Personas function as data-derived phenotypes or endotypes with differential treatment responses. In this way, NetraAI can support companion diagnostic development: if it identifies a subgroup defined by a biomarker that responds well, that biomarker can then be developed into a test for selecting patients.

The simplicity of NetraAI’s Personas—consisting of 2–4 variables—is intentional to ensure clinical feasibility. It is not practical for a physician to measure 20 different variables to determine if an individual fits within a subgroup; however, measuring a few key biomarkers or scores is relatively feasible. By ensuring that Personas are compact, NetraAI’s outputs are ready for translation into clinical decision support rules. A handful of tests (e.g., a lab test and a questionnaire that define Persona A) can triage patients, enabling real-time decision support to match a new patient to a likely responder Persona.

Ultimately, NetraAI doesn’t just predict what will happen—it suggests why it might happen by providing feature combinations. This can spur further scientific investigation: researchers might conduct laboratory experiments to understand the biology linking the features in a discovered Persona to outcomes.

\section{Broader AI Research Impact: Toward Collaborative AI Systems}

The meta-evolutionary design—NetraAI as experimentalist model guided by an LLM theorist—offers a glimpse into a broader paradigm of AI design: collaborative AI systems where different specialized AI components communicate, refine, and audit one another. In our case, the LLM provides a kind of global reasoning and reflection, while the core learning engine provides local learning and precision.

The LLM Strategist develops a sort of “theory-of-mind” about NetraAI’s learning process, monitoring what NetraAI is doing and adapting its strategy accordingly. This is reminiscent of how humans supervise and guide ML, but here an AI is doing that job, making the system more introspective—NetraAI is not blindly optimizing a loss, but is part of a dialog where its choices are examined and adjusted by another AI agent with a broader perspective.

Importantly, this architecture can be generalized to other domains; one can imagine a scientific discovery platform where an LLM proposes experiments and interprets results, while various specialist models (e.g., for chemistry, genomics, or imaging) do the heavy lifting on data and feedback.

This could lead to AI systems that can self-monitor and self-explain by design: when unexpected results arise, the strategist LLM can articulate hypotheses in natural language, rather than leaving us to interpret a black-box model post hoc, while NetraAI validates them quantitatively. This symbiosis paves the way for agentic platforms in science and medicine, combining grounded discovery with strategic reasoning.

\begin{figure*}[t]
\centering
\includegraphics[width=0.95\textwidth]{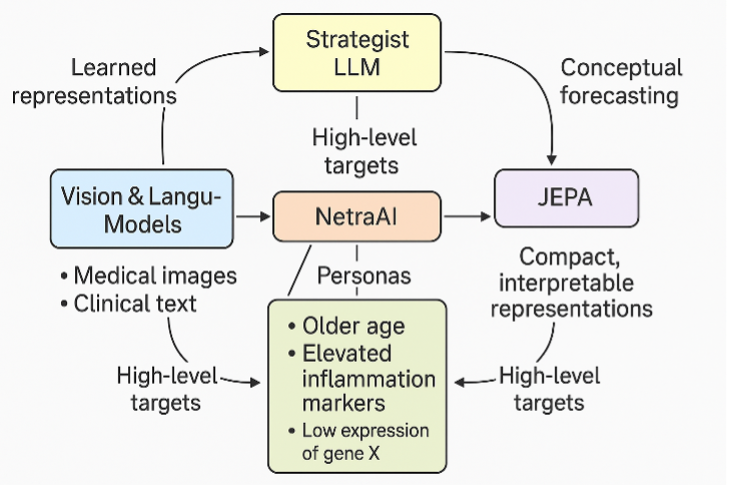}
\caption{Proposed depiction of the integration of NetraAI, LLM Strategist, foundational models, and JEPA models to develop an agentic AI system.}
\label{fig:agentic-architecture}
\end{figure*}

\subsection{Convergence of Disciplines}

NetraAI marries concepts from dynamical systems, evolutionary biology, meta-learning, interpretable ML, and collaboration with human expertise, producing a framework that resonates across fields—clinicians see validated patient cohorts, physicists recognize fixed-point attractors, biologists appreciate evolutionary feature selection, and AI researchers find parallels to meta-optimization.

This cross-disciplinary convergence positions NetraAI not merely as a predictive tool, but as a scientific partner that explains, guides, and evolves with the frontiers of scientific research.

\subsection{Toward Agentic AI: Fusion of NetraAI and Foundation Models}

NetraAI’s symbiosis with an LLM Strategist exemplifies how combining different AI species yields greater-than-sum capabilities. While NetraAI is already built for multimodal learning—handling structured clinical, behavioral, imaging-derived, and genomic data—its integration with foundation models introduces a new strategic layer.

Foundation models can extract latent signals from unstructured sources (e.g., radiology images, clinical notes, or biomedical literature), which can then be encoded into NetraAI’s feature space (e.g., “MRI phenotype A,” “textual mention of symptom X”) and statistically tested in NetraAI’s outcome-driven discovery pipeline. Conversely, NetraAI’s robust contraction dynamics and evolutionary feature search filter and validate only those patterns that persist across pseudo-temporal learning and meet outcome separation thresholds.

The complementary architecture—NetraAI as the grounded experimentalist and LLM for reasoning and interpretation—forms a self-improving ecosystem. As it analyzes more datasets, NetraAI generates high-signal Persona tokens, compact abstractions of variable bundles and outcome alignments, that can fine-tune the LLM Strategist, which in turn refines NetraAI’s exploration trajectory (e.g., seeding variable orders, suggesting termination criteria). Over time, this co-evolution amplifies both context-driven reasoning and structure-driven discovery.

Looking ahead, Persona tokens may underpin JEPA-style architectures, enabling models to predict conceptual dynamics such as transitions between patient subgroups over time (e.g., “Persona C to Persona A post-treatment”). This would allow for abstract, clinically meaningful prediction that aligns with emerging architectures prioritizing latent representation over surface-level pattern completion.

While strategic challenges remain—prompt alignment across models, mitigation of LLM hallucinations, coordination of compute resources across components—these can be managed through invoking foundation models at decision points, using distilled variants for routine tasks, and validating outputs through NetraAI’s statistical rigor.

Collectively, integrating NetraAI with foundational models and JEPA-style abstractions moves beyond monolithic AI toward a powerful agentic, layered AI system: one that specializes, communicates, and adapts, delivering interpretable, outcome-focused insights purpose-built for scientific discovery and clinical impact.

\section{The AI Modularity Hypothesis}

As artificial intelligence systems become increasingly complex, a new architectural paradigm is emerging: the \textbf{AI Modularity Hypothesis}. This hypothesis posits that future progress in AI will be driven not by scaling monolithic models, but by orchestrating ensembles of specialized, interoperable modules. Just as biological intelligence is composed of distinct systems—vision, language, memory—AI may achieve robustness, flexibility, and interpretability by mirroring this modular structure.

In this framework, different AI components are optimized for complementary roles: large language models (LLMs) serve as strategists and synthesizers, integrating unstructured information and offering contextual reasoning; vision and speech models handle modality-specific perception; and systems like NetraAI act as domain-specialized experimenters, extracting rigorous, interpretable insights from structured data. The coordination among these modules forms an agentic ecosystem, where generalists and specialists cooperate in dynamic workflows, much like interdisciplinary scientific teams.

NetraAI’s role within this ecosystem is emblematic of the modularity hypothesis. It does not attempt to do everything. Instead, it excels at discovering explainable patient subgroups in sparse, noisy clinical data—an area where foundation models typically underperform. Its ability to communicate through concept-level tokens (\textit{Personas}) makes it naturally compatible with other modules, such as JEPA-based predictors or LLMs trained for clinical summarization. By treating each module as a self-contained system with clear interfaces, the modular AI architecture becomes not only more interpretable but also more adaptable to new problems and data modalities.

This hypothesis shifts the design ethos of AI from “scaling everything” to “scaling coordination.” Instead of building a single model to solve every task, we build interoperable agents that excel at their respective roles and communicate through shared abstractions. NetraAI’s demonstrated ability to integrate with LLMs while preserving its own interpretive clarity is a step toward realizing this vision of intelligent modularity in clinical discovery.

\section{Conclusion}

NetraAI represents a turning point in the evolution of AI—from monolithic, data-hungry architectures toward domain-specialized, interpretable, and adaptive systems. Rather than relying solely on probabilistic pattern recognition, NetraAI leverages contraction-based dynamics, evolutionary feature search, and long-range memory to isolate and explain clinically meaningful subpopulations with high precision.

We have formalized NetraAI’s design, grounded in mathematical principles and tailored for clinical use. By uniting a dynamical learner with a strategist LLM, NetraAI forms a two-tiered meta-evolutionary loop that refines both its outputs and its own strategy over time. The result is a set of explainable Personas—compact feature bundles linked to distinct outcomes, surfaced with statistical confidence and narrative clarity—aligning with the goals of precision medicine while satisfying regulatory demands for transparency and fairness.

Rather than opposing large foundation models, NetraAI complements them: it channels general-purpose reasoning into focused, outcome-driven discovery. The integration of NetraAI with LLMs exemplifies a broader AI paradigm shift toward hybrid ecosystems, where specialized components—LLMs for breadth and linguistic fluency, NetraAI for structural depth, interpretability, and clinical alignment—cooperate to tackle high-stakes challenges like clinical trials.

\subsection*{Future Directions}

Looking forward, we envision several paths for advancement:

\begin{itemize}
    \item \textbf{Longitudinal Application:} Applying NetraAI to longitudinal data would enable continuous learning in adaptive trials or clinical monitoring settings.
    \item \textbf{Mechanistic Integration:} Linking Persona definitions to causal mechanistic biological models would strengthen insights.
    \item \textbf{Multi-strategist Architecture:} Expanding the strategist into a consortium of foundation models—each focused on clinical, statistical, or regulatory reasoning—would provide richer guidance and validation.
\end{itemize}

Ongoing collaborations are being pursued in oncology and psychiatry to validate NetraAI’s ability to reveal responder phenotypes that have real-world therapeutic implications. Ultimately, NetraAI embodies the future of AI systems that are not only intelligent, but also reflective, collaborative, and trustworthy. It captures a future in which learning is paired with explanation, scale is paired with strategy, and AI becomes an active partner in human discovery. This speciation of AI—across language, structure, memory, and reasoning—is not a detour, but a necessity for the future of scientific discovery and clinical practice.

\section*{Disclosures}

J.G, B.Q, M.T, C.C, and P.L were employed by NetraMark Corp. J.G declares that he owns substantial shares in NetraMark Holdings, which funded a major portion of this study.

L.P disclosures (Last 2 years): AbbVie, USA; Acumen, USA; Aicure, USA; Alexion, Italy; BCG, Switzerland; Astra-Zeneca, Italy; Boehringer Ingelheim International GmbH, Germany; EDRA-LSWR Publishing Company, Italy; GH-Pharma, Ireland; GLG-Institute, USA; Immunogen, USA; Johnson \& Johnson USA; LB-Pharmaceuticals, USA; Magdalena BioSciences, USA; MSD, Italy; Sanofi-Aventis-Genzyme, France and USA; Lundbeck, Denmark and Italy; NapoPharma, USA and EU; NetraMark, Canada; Pfizer Global, USA; Relmada Therapeutics, USA; Takeda, USA. Shares/Options: Relmada, NetraMark.

\onecolumn
\section*{References}
\begin{enumerate}
    \item Demis Hassabis – Nobel Prize lecture - NobelPrize.org. \url{https://www.nobelprize.org/prizes/chemistry/2024/hassabis/lecture/}.
    \item deepseek-ai/DeepSeek-V3 · Hugging Face. \url{https://huggingface.co/deepseek-ai/DeepSeek-V3}.
    \item Rudin, C. Stop explaining black box machine learning models for high stakes decisions and use interpretable models instead. \textit{Nat Mach Intell} \textbf{1}, 206–215 (2019).
    \item Lecun, Y. \textit{A Path Towards Autonomous Machine Intelligence}. (2022).
    \item Mannan, Md. A. et al. A Study of Banach Fixed Point Theorem and Its Applications. \textit{American Journal of Computational Mathematics} \textbf{11}, 157–174 (2021).
    \item Arora, S. \& Barak, B. \textit{Computational Complexity: A Modern Approach}. \url{https://theory.cs.princeton.edu/complexity/}.
    \item Nicolau, M., Levine, A. J. \& Carlsson, G. Topology based data analysis identifies a subgroup of breast cancers with a unique mutational profile and excellent survival. \textit{Proc Natl Acad Sci U S A} \textbf{108}, 7265–7270 (2011).
    \item Barnsley, M. F. \& Demko, S. G. \textit{Chaotic Dynamics and Fractals}. (2015).
    \item Singh, A. \textit{Mastering Feature Extraction: PCA, t-SNE, and LDA in Machine Learning Part-2}. \url{https://medium.com/@abhaysingh71711/mastering-feature-extraction-pca-t-sne-and-lda-in-machine-learning-part-2-5bb4f8ab0f9c}.
    \item Denardo, E. V. Contraction Mappings in the Theory Underlying Dynamic Programming. \textit{SIAM Review} \textbf{9}, 165–177 (1967).
    \item Goodfellow, I., Bengio, Y., Courville, A. \textit{Deep Learning}. (2016).
    \item Gao, Y. et al. When Raw Data Prevails: Are Large Language Model Embeddings Effective in Numerical Data Representation for Medical Machine Learning Applications? (2024).
    \item Qi, B. et al. Large Language Models as Biomedical Hypothesis Generators: A Comprehensive Evaluation. (2024).
    \item Lieberman, J. A. et al. Effectiveness of Antipsychotic Drugs in Patients with Chronic Schizophrenia. \textit{New England Journal of Medicine} \textbf{353}, 1209–1223 (2005).
    \item Géron, A. \textit{Hands-on Machine Learning with Scikit-Learn, Keras, and TensorFlow}. (2019).
    \item Lam, R. W. et al. Discovering biomarkers for antidepressant response: Protocol from the Canadian biomarker integration network in depression (CAN-BIND) and clinical characteristics of the first patient cohort. \textit{BMC Psychiatry} \textbf{16}, (2016).
    \item Aung, K. L. et al. Genomics-Driven Precision Medicine for Advanced Pancreatic Cancer: Early Results from the COMPASS Trial. \textit{Clin Cancer Res} \textbf{24}, 1344–1354 (2018).
\end{enumerate}

\section*{Abbreviations}

\begin{tabular}{ll}
AI & artificial intelligence \\
BCE & binary cross-entropy \\
CR & complete response \\
CRP & C-reactive protein \\
DNN & deep neural network \\
FFX & FOLFIRINOX \\
GnP & gemcitabine + nab-paclitaxel \\
LLM & large language model \\
MDD & major depressive disorder \\
MoE & Mixture-of-Experts \\
PCOF & perphenazine completed, olanzapine failed \\
PDAC & pancreatic ductal adenocarcinoma \\
PFOC & perphenazine failed, olanzapine completed \\
PR & partial response \\
SD & stable disease \\
\end{tabular}

\section*{License}

This work is licensed under a \textbf{Creative Commons Attribution–NonCommercial–NoDerivatives 4.0 International License}.

You may share this work with attribution for non-commercial purposes, but you may not modify or reuse it for commercial applications without explicit permission from the authors.

\end{document}